\documentclass[10pt,twocolumn,letterpaper]{article}

\usepackage{cvpr}              
\definecolor{cvprblue}{rgb}{0.21,0.49,0.74}
\usepackage[pagebackref,breaklinks,colorlinks,allcolors=cvprblue]{hyperref}

\def\paperID{*****} 
\def\confName{CVPR}
\def\confYear{2026}

\title{Synthetic Data Generation for Long‑Tail Medical Image Classification: A Case Study in Skin Lesions}

\author{Jiaxiang Jiang\\
Intel\\
{\tt\small jiaxiang.jiang@intel.com}
\and
Mahesh Subedar\\
Google\\
{\tt\small msubedar@gmail.com}
\and
Omesh Tickoo\\
Intel\\
{\tt\small omesh.tickoo@intel.com}
}

\begin{document}
\maketitle
\begin{abstract}
    
    Long‑tailed class distributions are pervasive in multi-class medical datasets and pose significant challenges for deep learning models which typically underperform on tail classes with limited samples. This limitation is particularly problematic in medical applications, where rare classes often correspond to severe or high‑risk diseases and therefore require high diagnostic accuracy. Existing solutions—including specialized architectures, rebalanced loss functions, and handcrafted data augmentation—offer only marginal improvements and struggle to scale due to their limited and largely deterministic variability.
    To address these challenges, we introduce a diffusion‑model‑driven synthetic data augmentation pipeline tailored for medical long‑tailed classification. Our approach features a novel inpainting diffusion model combined with an Out‑of‑Distribution (OOD) post‑selection mechanism to ensure diverse, realistic, and clinically meaningful synthetic samples. Evaluated on the ISIC2019 skin lesion classification dataset, one of the largest and most imbalanced medical imaging benchmarks, our method yields substantial improvements in overall performance, with particularly pronounced gains on tail classes with more than $28\%$ improvement on the class with the fewest samples. These results demonstrate the effectiveness of diffusion-based augmentation in mitigating long‑tail imbalance and enhancing medical classification robustness.
\end{abstract}
\section{Introduction}

Despite the impressive progress brought by modern deep neural network (DNN) architectures \cite{resnet,tan2019efficientnet,dosovitskiy2020image,liu2022convnet_convnext}, image classification performance in conventional benchmarks is approaching saturation. However, these gains rely heavily on large, balanced datasets, a condition rarely met in real‑world medical imaging. Medical images like dermoscopy inherently exhibit long‑tailed label distributions, where a few dominant “head” classes contain most samples, while numerous clinically important “tail” classes are severely under‑represented. Models trained on such skewed data often achieve competitive overall accuracy yet fail to generalize to tail classes, which is particularly concerning because these classes may correspond to rare but high‑risk diseases. Consequently, improving recognition performance under long‑tailed distributions has become a critical challenge, and long‑tailed medical image classification (LTMIC) has emerged as an important research direction within the computer vision community.

Existing solutions for LTMIC can be grouped into three categories \cite{ju2024monica,LT_survey_1}:
loss function design, model improvement, and class re-balancing.
Cross-entropy loss is the most widely used loss function in deep learning model training. 
However, in LTMIC setting, this loss ignores the class imbalance in number of samples for each class. 
Therefore, variations of training loss functions are designed to capture the intrinsic imbalance in the training set \cite{cao2019learning_LT_loss,hong2021disentangling_LT_loss2,kini2021label_LT_loss3,lin2017focal_LT_loss4,ren2020balanced_LT_loss5,tan2020equalization_LT_loss6}. 
Instead of adjusting training loss function directly, \cite{zhang2021distribution_logits_loss7} proposes to adjust the prediction logits of a neural network model, which uses an calibration function to adjust the logits.
The calibration function is learnt by matching the calibrated prediction to some pre-defined more balanced class distribution.

Model improvement methods design specific model architectures to solve LTMIC. 
Most LTMIC model can be divided into two stages, feature representation leaning and classifier. 
Therefore, some methods aim to design better learning representation model \cite{kang2020exploring_feature_learning1,zhang2017range_feature_learning2} and another methods are mainly to design suitable classifier \cite{foret2020sharpness_class1,li2021self_class2,zhou2020bbn_class3,ISIC2019class_main_result_comp1,isic_ext_result_comp2}.
\cite{kang2020exploring_feature_learning1} learns discriminative features by using the overall distance among all samples pairs within one training batch and \cite{zhang2017range_feature_learning2} enlarges the inter-class distance and minimize the largest distance between intra-class samples in the feature space.
DataFuse\cite{isic_ext_result_comp2} shows introducing external data can result in a better feature representation.
\cite{zhou2020bbn_class3} proposes to design the classifier with two branches, one is a conventional learning branch and another is a re-balancing branch, to target the imbalance of the training set.
\cite{li2021self_class2} explores distillation technique to design the long-tail classifier.
\cite{zhou2020bbn_class3} defines a loss sharpness in addition to loss value itself as the objective function to mitigate the imbalance problem.
MRE\cite{ISIC2019class_main_result_comp1}, MME\cite{zhou2019multi_comp3}, MME\cite{tsai2023skin_style_comp5} ensemble individual classifiers in a multi-expert framework. 
Empirically, these ensembling methods achieve the state-of-the-art performance mainly by reducing the model uncertainties to obtain robust predictions.
However, all model improvement methods and special loss function design methods share a common a commonality: they acknowledge the data imbalance and focus on the training of models \cite{diffLT}.
They both require meticulous design and thus are too sensitive to hyperparameters and have limited generalization capability. 
For example, one carefully designed loss function works well on one model but fails on another model. 
Also, different ensembles of models can have quite different results even on the same dataset.

Another group of methods are class-rebalancing methods which aim to re-balance the negative influence of the imbalance of the training set by adjusting data distribution. One of the most studied methods in this group are re-sampling based methods \cite{chawla2002smote_resample_intro1,zhang2021learning_resampling_intro2,kang2019decoupling_resampling_intro3} which have been explored to re-balance classes by adjusting the number of samples per class in each sample batch for model training.
Instead of simply resampling the training set, recent work \cite{ahn2023cuda} develops a class specific data augmentation method to adjust the data set distribution.
\cite{ahn2023cuda} shows that adding more samples to adjust the distribution can increase average classification performance as well as class-wise performance.
However, the augmentation method proposed in \cite{ahn2023cuda} is too simple and it does not capture distribution of each class .

Recent development of image generation techniques \cite{ho2020denoising_ddpm,song2020denoising_ddim,robin2022high_ldm} show great potential of generating high quality natural images.
Those methods utilize diffusion models \cite{song2020denoising_ddim,ho2020denoising_ddpm} to produce visually appealing images both with or without conditions. 
There is a large amount of following work \cite{blattmann2023stable_diffusion,saharia2022photorealistic_imagen,zhang2023adding_controlnet} in pushing the limits of generating better photorealistic images with high FID and PSNR scores.
Despite its success in generating visually appealing images for digital design, applications of diffusion models in limited data scenarios especially domain specific scenarios remain under-explored.

\cite{CBDM_result_com4} and \cite{diffLT} explore how to make diffusion models to generate images for the long-tailed dataset. 
\cite{CBDM_result_com4} proposes a class-balancing diffusion model (CBDM) to generate diverse and good quality images for each class in the long-tailed datasets.
CBDM\cite{CBDM_result_com4} demonstrate its diverse generation capability especially for tail classes.
Based on CBDM \cite{CBDM_result_com4}, \cite{diffLT} show that not all samples are equally important in training diffusion models for long-tailed datasets.
\cite{diffLT} show that approximately-in-distribution (AID) samples play more important role than in-distribution (ID) and out-of-distribution (OOD) samples.
However, both \cite{CBDM_result_com4} and \cite{diffLT} show great performance on natural image long tail datasets where the distances between different class distributions are relatively large than medical long tail datasets such as dermatoscope imagery.

\begin{figure*}[!htb]
     \centering
     \includegraphics[width=1\textwidth]{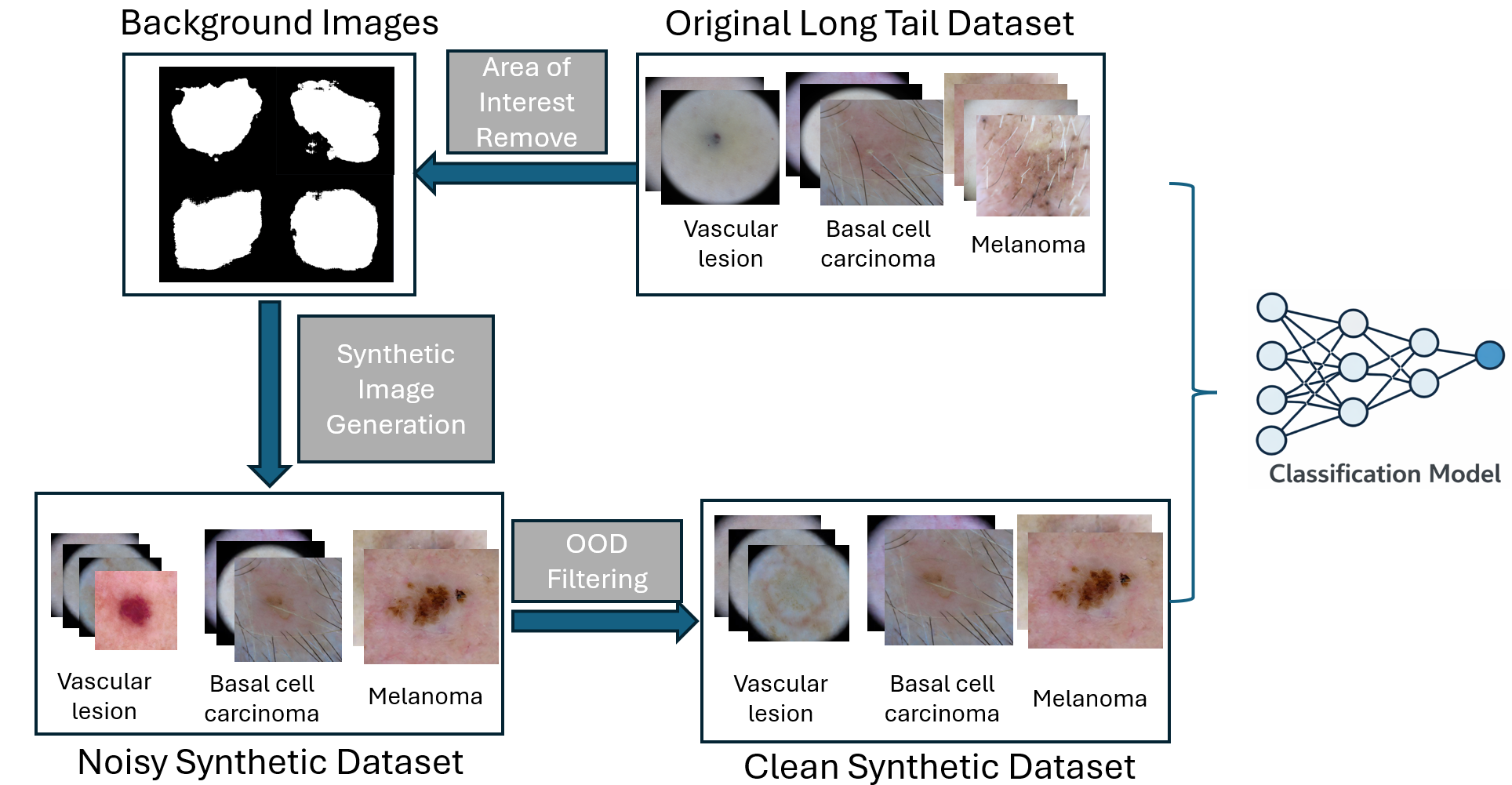}

  \caption{
  The overal pipeline of our synthetic data generation approach. First, a segmentation method is used to get background images of the original long tail dataset. Next, a finetuned inpaint diffusion model is used to generate noisy synthetic images which can contain OOD image samples. Then an OOD filtering is applied to noisy synthetic dataset to get clean synthetic dataset. Last, the clean synthetic dataset is combined with the original long-tail dataset to train a final classification model.  
  }
  \label{fig:method_overview}
\end{figure*}

We propose a synthetic data generation approach to solve the Long-Tail Medical Image Classification (LTMIC) problem, as illustrated in Fig \ref{fig:method_overview}. First, we extract the area of interest from dermatoscopy images using the method described in \cite{ma2024segment_medsam}. Subsequently, we generate a set of synthetic images using our fine-tuned inpainting diffusion model. Specific details regarding the diffusion model architecture and fine-tuning techniques are described in section \ref{sec:method}.

A medical Out-of-Distribution (OOD) method \cite{cai2025medianomaly} is utilized to filter out OOD samples for each class. Following this filtering process, we combine the synthetic samples with the original dataset to train a classifier. We demonstrate that the inclusion of synthetic samples significantly improves classification results, particularly for the tail classes. Since this paper introduces a novel synthetic data generation approach, we utilize only a simple CNN-based classifier \cite{ISIC2019class_main_result_comp1} in our experiments without further data augmentation or model ensembles.

In summary, the main contributions of this work are as follows: 
\begin{enumerate}
    \item We propose a novel synthetic data generation pipeline incorporating an inpainting diffusion model and OOD filtering to address the LTMIC problem;
    \item We introduce a novel inpainting diffusion model architecture specifically designed for medical image generation;
    \item We present a novel OOD-based sampling strategy to fine-tune diffusion models on medical long-tail datasets.
\end{enumerate}

\section{Approach} \label{sec:method}
\begin{figure*}[!htb]
     \centering
     \includegraphics[width=1\textwidth]{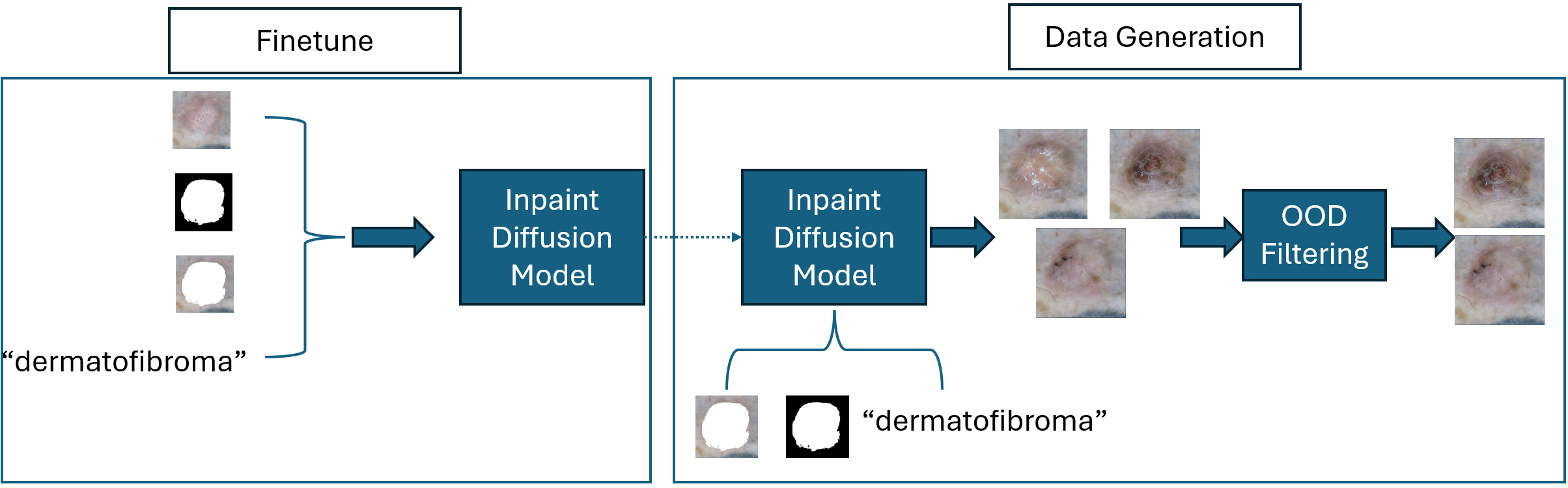}

  \caption{
  Inpaint diffusion model finetuning and data generation. First, we finetune an inpainting diffusion model using the original long tailed training dataset, their associated binary area of interest masks from a segmentation model, background of original dermoscopy images, and their class labels. Next, the finetuned diffusion model is used to generate synthetic samples. The finetuned model takes input the background images of the original training set images and their associated class labels. Finally, an OOD method is used to filter out OOD samples to make the synthetic dataset cleaner to the classification model.
  }
  \label{fig:data_gen_overview}
\end{figure*}

\subsection{Diffusion Model Background}

In the standard diffusion model \cite{ho2020denoising_ddpm}, during training, the forward process is characterized by the following equation with a batch of input images $\textbf{x}_0$.
\begin{equation}
    q(\textbf{x}_t|\textbf{x}_0) = \mathcal{N}(\sqrt{\bar{\alpha_t}}\textbf{x}_0,(1-\bar{\alpha_t}\textbf{I}))
\end{equation}
where $\bar{\alpha_t}=\prod_{i=1}^{t}(1-\beta_i)$ is calculated through pre-defined variance scheduler $\{\beta_t\in(0,1)\}_{t=1}^T$.
Next, a diffusion model parameterized by parameter $\theta$ to get the posterior probability distribution $p_{\theta}(\textbf{x}_{t-1}|\textbf{x}_t,t)$.
Instead of modeling this posterior probablity directly, \cite{robin2022high_ldm} uses a UNet\cite{ronneberger2015u_unet} with parameter $\theta _d$ to model $\epsilon_t$ which is the noise added to the original image to get $\textbf{x}_t$.
\cite{ruiz2023dreambooth} shows that the diffusion model parameter $\theta_d$ can be finetuned with few images with the following objective function to get high quality images:
\begin{equation}
    L=\mathbb{E}_{t\sim[1,T]}[||\epsilon _t-\epsilon _{\theta_d}(\sqrt{\tilde{\alpha}_t}\textbf{x}_0+\sqrt{1-\tilde{\alpha}_t}\epsilon_t,t,y_{emb})||^2]
\end{equation}
where $\epsilon_{\theta_d}$ is the noise estimated by the UNet with parameter $\theta_d$ and $y_{emb}$ is the embedding of the image class label $y$.
During reverse process, noise is estimated by the finetuned UNet at each time step $t$, and new sample is generated by subtracting estimated $\epsilon_t$ iteratively from pure noise sample $\textbf{x}_T\sim\mathcal{N}(0,I)$.

\subsection{Method Overview}
The overal data generation pipeline is illustrated in Fig~\ref{fig:method_overview}. 
Mathematically, for general image classification dataset with $N$ number of data samples and a set of classes $C$, $\mathcal{D_{LT}}=\{(x_i,y_i)\}_{i=1}^{N}$ where $x_i$ represents $i_{th}$ image and $y_i \in C$ represents $i_{th}$ image label.
In the long-tail setting, $|c_1|\geq|c_2|\geq\dots\geq|c_M|$ for each $c_j\in C$, $M$ represents total number of different classes of the dataset, $|c_j|$ represents number of samples in $c_j$, and $M$ represents total number of classes in $C$.
In this paper, we create a synthetic $\mathcal{D}_{syn}$ which has the same set of classes $C$ as $\mathcal{D_{LT}}$,
For each class $c^{syn}_{j}$ in $\mathcal{D}_{syn}$, $|c^{syn}_{j}| = max\{0,|c_1|-|c_j|\}$.
Because of stochasticity in diffusion model-based data generation approach, it is not guaranteed that all samples in $\mathcal{D}_{syn}$ are good samples to train the downstream classification model.
Thus $\mathcal{D}_{syn}$ is a noisy synthetic dataset and an OOD method is used to filter out noisy samples to get clean synthetic dataset $\mathcal{D}_{syn}^{clean}$.
For $j_{th}$ class $c_{j}^{clean}$ in $\mathcal{D}_{syn}^{clean}$, $|c_{j}^{clean}|=\gamma \times |c^{syn}_{j}|$ where $\gamma \in[0,1]$ is a hyperparameter and it represents the percentage of clean samples in each class in $\mathcal{D}_{syn}$.
Last, the augmented dataset $\mathcal{D}_{syn}^{clean}\cup\mathcal{D_{LT}}$ is used to train the classification model from \cite{ISIC2019class_main_result_comp1}.

Since the main innovations of this paper are focusing on data generation part of the pipeline so this section will mainly describe how to acquire inpaint diffusion model and use it to generate synthetic samples to train the classification model.
As illustrated in fig \ref{fig:data_gen_overview}, an inpaint diffusion model is finetuned with the original dermoscopy image $x_i$, binary area of interest mask $b_i$, background of original dermoscopy image $d_i$, and their class label $y_i$.
The finetune process makes the inpaint diffusion model learn the distribution of masked out region conditioned on $d_i$ and $y_i$.
After finetuning, the inpaint is used to generate images following the standard reverse process as described in \cite{robin2022high_ldm,CBDM_result_com4,diffLT}.
\subsection{Proposed Inpaint Diffusion Model}
\subsubsection{Inpaint Diffusion Model Architecture}
\begin{figure*}[!htb]
     \centering
     \includegraphics[width=1\textwidth]{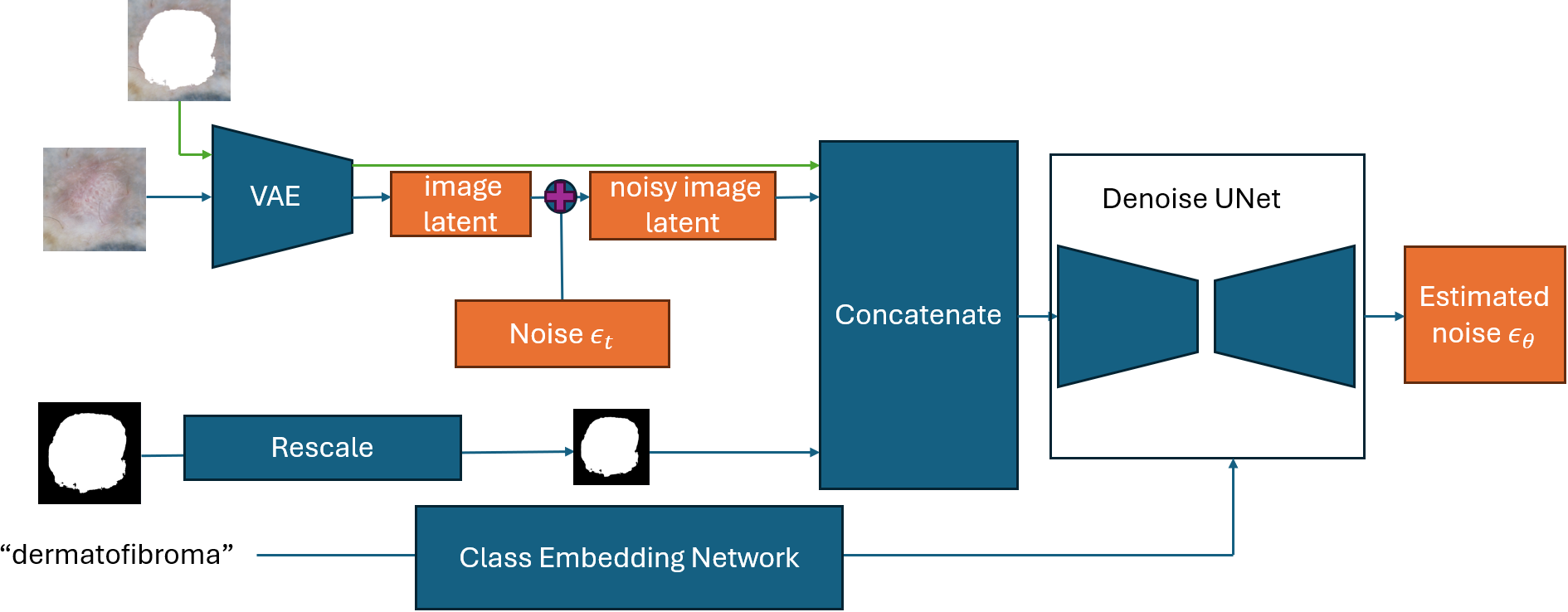}

  \caption{
  Inpaint diffusion model architecture.
  }
  \label{fig:diffusion_model}
\end{figure*}
Our inpaint diffusion model is illustrated in Fig~\ref{fig:diffusion_model}.
During finetuning, the model takes input of dermoscopy image $\textbf{x}$, binary segmentation image of area of interest $\textbf{b}$, dermoscopy background image $\textbf{d}$, and class label $\textbf{y}$.
One of the main challenges in multi-class domain specific image generation for image classification is to generate images distinguishable enough between different classes while also within the real distribution of label $\textbf{y}$.
The general text embedding models are not suitable for highly specialized description of skin diseases. Hence, we train a simple fully connected class embedding network as shown in Fig~\ref{fig:diffusion_model}.
The class embedding network is parameterized by learnable parameters $\theta_c$.
Both original image $\textbf{x}$ and dermoscopy background image $\textbf{d}$ are encoded into latent space by a pre-trained Variational Auto Encoder (VAE) and noise $\epsilon_t$ is added to get noisy latents $\textbf{z}_{xt}$ and $\textbf{z}_{dt}$ respectively at diffusion time step $t$.
The binary mask $\textbf{b}$ is rescaled as $\textbf{b}^\ast$ to match the shape of noisy latents $\textbf{z}_{xt}$ and $\textbf{z}_{dt}$.
$\textbf{z}_{xt}$, $\textbf{z}_{dt}$, and $\textbf{b}^\ast$ are concatenated into noisy latent $\textbf{z}=[\textbf{z}_{xt},\textbf{z}_{dt},\textbf{b}^\ast]$ as the input to the diffusion UNet model parameterized by $\theta_d$.
Therefore, the total learnable parameters include $\theta = \theta_d \cup \theta_c$. 
Thus the funetune objective function becomes:
\begin{equation}
    L=\mathbb{E}_{t\sim[1,T]}[||\epsilon _t-\epsilon _{\theta}(\textbf{z},t,y_{emb})||^2]
\end{equation}

\subsubsection{Finetuning with sample selection}
In order to make the diffusion model learn the distribution of each class equally well, we apply an oversampling strategy to make sure the number of samples of each class is the same.
The idea is to assign a weight $w_i^j$ for $i_th$ sample in class $c_j$.
Then we repeat samples in class $c_j, j=2,3,\dots,M$ based on $w_i^j$ to make sure the number of finetuning samples for each class is equal to $|c_1|$.
Next, we will describe how we compute sample weights $w_i^j$. 

As \cite{diffLT} suggests that different samples can have different importance in training the generative model and approximately-in-distribution (AID) samples, which locate slightly far from the center of the distribution play the most important role.
Therefore, we utilize the self-supervised medical anomaly detection method \cite{cai2025medianomaly} to train for each class $c_j, j=2,3,\dots,M$ using images $\textbf{x}$ from $\mathcal{D_{LT}}$. 
After training, for each class $c_j$, we compute anomaly score $\mathcal{A}_i^j$ for each image $\textbf{x}_i^j$ in this class.
Then weight $w_i^j$ is computed as
\begin{equation} \label{equa:weights}
    w_i^j = \frac{1}{e^{|\bar{\mathcal{A}}^j-\mathcal{A}_i^j|}}
\end{equation}
where $\bar{A}^j$ is the median value of $\bar{A}_i^j, i=1,2,\dots|c_j|$. 
$|\bar{\mathcal{A}_i^j}-\mathcal{A}_i^j|$ is the distance measure of how far sample $i$ is from AID samples. The samples that are closer in distance are given higher sampling probability during the finetuning step.

\section{Experimental Results}
\begin{figure*}[!htb]
     \centering
     \includegraphics[width=0.55\textwidth]{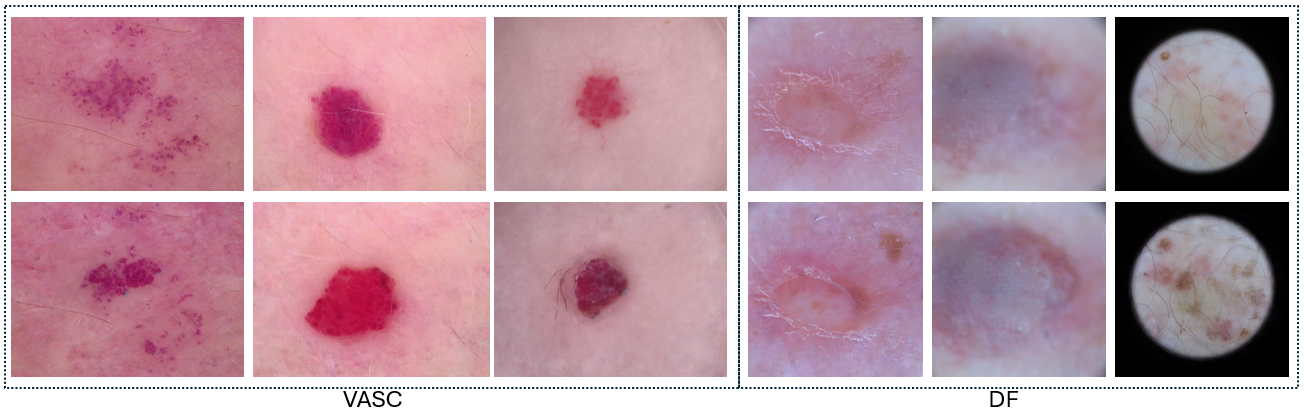}

  \caption{
  Example images (top row) from the dataset and the corresponding synthetic images (bottom row) from our inpaint diffusion model. Images are selected from two classes with the least number of samples: VASC and DF. 
  }
  \label{fig:results}
\end{figure*}

ISIC2019 skin lesion dataset is used to test our method.  
The dataset consists 25331 images of 8 classes: Melanoma (MEL), Melanocytic Nevus (NV), Basal Cell Carcinoma (BCC), Actinic Keratosis (AKIEC), Benign Keratosis (BKL), Dermatofibroma (DF), Vascular Lesion (VASC), and Squamous Cell Carcinoma (SCC).
The number of images for each class vary significantly: MEL images is 4,522; NV is 12,875; BCC is 3,323; AKIEC is 867; BKL is 2,624; DF is 239; VASC is 253; and SCC is 628. Fig.\ref{fig:results} shows example images from two tail classes DF and VASC, and synthetic images from the corresponding input image. 

The 5 fold cross validation is used for the experiments. 
We split the dataset into 5 equal folds and each class is split proportionally into 5 equal folds.
For each round, we use 4 folds to finetune our generative model, generate synthetic samples, and is then combined with the clean synthetic dataset to train a classification model.
The remaining 1 fold is used to get the test results from the trained classification model.
We iterative this process five times to get the average performance of the classification models.


To evaluate our approach across all eight classes, we utilize the four metrics defined in the ISIC 2019 challenge: Balanced Multiclass Accuracy (BMA), sensitivity, specificity, and F1 score. In this multiclass context, these metrics are derived from the classification confusion matrix. Specifically, BMA is calculated as the mean recall across all classes (i.e., the average of the diagonal elements divided by the total positive instances per class). All reported metrics represent the average results across 5-fold cross-validation.

Table~\ref{tab:results} compares our approach against current SOTA methods. For a fair comparison, we utilize the single classification model from \cite{ISIC2019class_main_result_comp1} without multi-resolution ensembling (MRE) (MRE Without Ensembling). Our synthetic data generation approach surpasses all multi-model ensembling methods \cite{ISIC2019class_main_result_comp1, zhou2019multi_comp3, tsai2023skin_style_comp5} across all four metrics by a significant margin. Notably, our method outperforms \cite{isic_ext_result_comp2} by approximately $4\%$ in BMA without requiring any external data. Among diffusion-based synthetic data approaches, our specific diffusion architecture and sampling strategy yield superior performance compared to \cite{ruiz2023dreambooth, CBDM_result_com4}. Furthermore, the results highlight that OOD filtering consistently improves performance across all evaluation metrics.

\begin{table*}[!htb]
    \caption{Benchmarking results for our methods against SOTA methods on ISIC2019 dataset. BMA, F1 score, sensitivity, and specificity are used to capture the performance of various methods.}
    \centering
    \begin{tabular}{|c|c|c|c|c|}
    \hline
\textbf{Methods} & \textbf{BMA} & \textbf{F1 score} & \textbf{sensitivity} & \textbf{specificity} \\ \hline    MRE\cite{ISIC2019class_main_result_comp1}    &0.757&0.726&0.693&0.957 \\ \hline
    CNNE\cite{zhou2019multi_comp3}& 0.753 & 0.700& 0.692& 0.944 \\ \hline
    DataFuse\cite{isic_ext_result_comp2} &0.762&0.674&0.690&0.951 \\ \hline
    MME\cite{tsai2023skin_style_comp5} &0.637&0.596&0.624&0.933 \\ \hline
    CBDM\cite{CBDM_result_com4}&0.739&0.706&0.739&0.957 \\ \hline
    Dreambooth\cite{ruiz2023dreambooth} &0.740	&0.719	&0.740	&0.957 \\ \hline
    MRE Without Ensembling  & 0.677	&0.700	&0.677	&0.956 \\ \hline
    Our Approach Without Filtering &0.800	&0.760	&0.800	&0.963 \\ \hline
    Our Approach &\textbf{0.802}	&\textbf{0.780}	&\textbf{0.802}	&\textbf{0.965} \\ \hline
         
    \end{tabular}
    
    \label{tab:results}
\end{table*}
\begin{table*}[!htb]
    \caption{Impact of Synthetic Data Augmentation on Per-Class Classification Accuracy.}
    \centering
    \begin{tabular}{|c|c|c|c|}
    \hline
\textbf{Skin Lesion Classes}     &  \textbf{MRE Without Ensembling} & \textbf{Our approach without filtering} & \textbf{Our approach}\\ \hline
    MEL   &0.694 &\textbf{0.759}	&0.7583 \\ \hline
    NV   & 0.753	&0.762	&\textbf{0.793} \\ \hline
    BCC &0.792 &	0.864	&\textbf{0.882} \\ \hline
    AKIEC &0.604	&0.709	&\textbf{0.724} \\ \hline
    BKL& 0.527	&0.655 &	\textbf{0.668}\\ \hline
    DF &0.611	&\textbf{0.897}	&0.880 \\ \hline
    VASC& 0.876	&\textbf{0.969}	&0.965 \\ \hline
    SCC &0.561	&\textbf{0.772}	&0.745 \\ \hline

    \end{tabular}
    
    \label{tab:results_perclass}
\end{table*}

\begin{figure*}[!htb]
     \centering
     \includegraphics[width=0.35\textwidth]{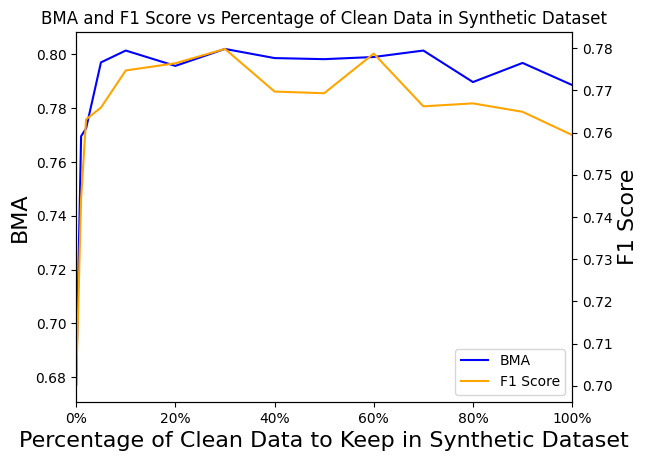}
     \includegraphics[width=0.35\textwidth]{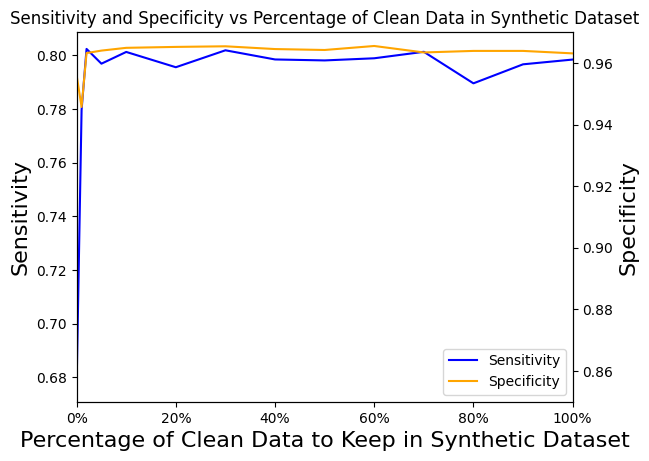}
       \caption{
  Ablation study on how $\gamma$, the percentage of clean samples affect the final classification performance. Specificity is the least sensitive metric to the hyperparameter. For other three evaluation metrics, the performance significantly improved with even a little amount of synthetic data. Then as $\gamma$ increases, the performance stays flat or decrease. }
  \label{fig:ablation}
\end{figure*}
We further analyze how synthetic data influences classification performance across individual classes using class-wise accuracy metrics, as shown in Table~\ref{tab:results_perclass}. We can clearly observe that classes with fewer original samples (tail classes) get great benefit from the synthetic data. For example, \textit{DF}—the tail class with the fewest samples—shows an accuracy improvement of over $28\%$, significantly exceeding the overall average performance gain.

Another key finding is that OOD filtering tends to improve head-class performance more than tail-class performance. This occurs because head classes already possess sufficient clean training data; consequently, synthetic samples without filtering introduce disproportionate noise for these classes. In contrast, tail classes benefit so substantially from the increased sample volume that they can tolerate higher levels of noise within the synthetic data than head classes can.

As mentioned above, OOD filtering plays a role in balancing head and tail classes classification accuracy. Thus, for the ablation study, we investigate how the hyperparameter $\gamma$, the percentage of clean samples that we keep in each class affects the final classification performance. As Fig.\ref{fig:ablation} shows, specificity almost does not change with $\gamma$. For other three evaluation metrics, as $\gamma$ increase from 0 to 1 (0 means use original training dataset only and 1 means no OOD filtering for synthetic dataset), the performance first get improved sharply and then those metrics stay flat or decrease. Based on Fig \ref{fig:ablation}, the best $\gamma$ value should be between $0.2\sim0.6$.

\section{Conclusion}


In this paper, we proposed a novel synthetic data generation approach to address the challenges of LTMIC, using skin lesion classification as a primary case study. Our method integrates a novel diffusion model with out-of-distribution (OOD) filtering to ensure the generation of robust, in-distribution synthetic samples. Experimental results demonstrate that fine-tuning downstream models with our synthetic data consistently outperforms current SOTA methods.

While our approach is model-agnostic and can be integrated with any architecture, treating data generation and classification as independent processes may be suboptimal. Therefore, our future work will focus on the joint optimization of these two components to further enhance LTMIC performance.

\clearpage
{
\bibliographystyle{ieeetr}
\bibliography{main}
}


\end{document}